\documentclass{article}

\usepackage{arxiv}

\usepackage[utf8]{inputenc} % allow utf-8 input
\usepackage[T1]{fontenc}    % use 8-bit T1 fonts
\usepackage{hyperref}       % hyperlinks
\usepackage{url}            % simple URL typesetting
\usepackage{booktabs}       % professional-quality tables
\usepackage{amsfonts}       % blackboard math symbols
\usepackage{amsmath,amsthm,amssymb}
\usepackage{nicefrac}       % compact symbols for 1/2, etc.
\usepackage{microtype}      % microtypography
\usepackage{cleveref}       % smart cross-referencing
\usepackage{lipsum}         % Can be removed after putting your text content
\usepackage{graphicx}
\usepackage[square,numbers]{natbib}
\usepackage{doi}

\usepackage{algpseudocode}
\usepackage{algorithm}
\usepackage{subcaption}
\usepackage{comment}
\usepackage{bm}
\usepackage{chngcntr}
\usepackage{bibunits}
\usepackage{multirow}
\newcommand{\E}[1]{\mathbb{E}\left[#1\right]}

\counterwithin*{theorem}{section}
\title{A Contextual Combinatorial Semi-Bandit  Approach to Network Bottleneck Identification}

\hypersetup{
pdftitle={A Contextual Combinatorial Semi-Bandit  Approach to Network Bottleneck Identification},
pdfsubject={cs.LG, stat.ML},
pdfauthor={Fazeleh Sadat Hoseini, Niklas {\AA}kerblom, Morteza Haghir Chehreghani},
pdfkeywords={Contextual Bandits, Combinatorial Bandits, Bottleneck Identification, Online Learning},
}

\begin{document}

\author{Fazeleh Hoseini \\
	Department of Computer Science and Engineering\\
	Chalmers University of Technology\\
	Gothenburg, Sweden \\
	\texttt{fazeleh@chalmers.se} \\
        \And
        Niklas {\AA}kerblom \\
	Volvo Car Corporation \\
	Gothenburg, Sweden \\[3pt] % add vspace for next institution
	Department of Computer Science and Engineering\\
	Chalmers University of Technology\\
	Gothenburg, Sweden \\
	\texttt{niklas.akerblom@chalmers.se} \\
        \And
	Morteza Haghir~Chehreghani \\
	Department of Computer Science and Engineering\\
	Chalmers University of Technology\\
	Gothenburg, Sweden \\
	\texttt{morteza.chehreghani@chalmers.se} \\
}

\maketitle 

\begin{abstract}
Bottleneck identification is a challenging task in network analysis, especially when the network is not fully specified. To address this task, we develop a unified online learning framework based on combinatorial semi-bandits that performs bottleneck identification in parallel with learning the specifications of the underlying network. Within this framework, we adapt and study various combinatorial semi-bandit methods such as epsilon-greedy, LinUCB, BayesUCB, NeuralUCB, and Thompson Sampling. In addition, our framework is capable of using contextual information in the form of contextual bandits. Finally, we evaluate our framework on the real-world application of road networks and demonstrate its effectiveness in different settings.
\end{abstract}


\section{Introduction} \label{sec:Introduction}

Bottleneck identification is an essential task in network analysis with numerous important applications, such as traffic planning and road network management. For example, in a road network, the road segment with the highest cost is described as a path-specific bottleneck on a path between a source node and a destination node. The cost or weight can be defined according to specific criteria, such as travel time, energy consumption, etc. The aim is to find a path which minimizes the bottleneck among all paths connecting the source and destination nodes.

Bottleneck identification can thus be characterized, in a given road network graph, as finding a path with the smallest maximum edge weight among the paths connecting the source node and the destination node, i.e., finding the \textit{minimax edge}. By negating the edge weights, bottleneck identification can also be viewed as the widest path problem or the maximum capacity path problem \cite{pollack1960maximum}. 
It is often assumed that the network parameters, such as the edge weights, are given in a classical minimax problem. However, in some real-world cases, the edge weights are stochastic and initially unknown, and have to be learned while solving the minimax problem. Thus, this problem can also be formulated a sequential decision-making problem under uncertainty.

The Multi-Armed Bandit (MAB) problem is a classical approach for modelling online decision-making problems.
In a MAB problem, an agent repeatedly performs an action (i.e., plays an \textit{arm}), receives a stochastic reward drawn from an unknown reward distribution, and updates estimates of the reward distribution parameters. In each iteration, the agent chooses an arm to play based on the observed rewards of the previously played arms.
The goal is to maximize the cumulative reward over a pre-determined time horizon by making a trade-off between exploring an environment to gain new knowledge and exploiting previously collected knowledge. Any \textit{greedy} method is based on pure exploitation, i.e., by always selecting actions which maximize the current estimates of the expected rewards. 
Towards more exploration, the $\epsilon_t$-greedy algorithm \cite{chen2013combinatorial} chooses uniformly random actions with probability $\epsilon_t$, decaying with each time step $t$, and otherwise exploits them greedily. If the horizon tends to infinity, the $\epsilon_t$-greedy algorithm guarantees that the expected reward estimate of each arm will converge to the actual expected value, since the agent is expected to play each arm infinitely many times. However, other algorithms make better use of the reward history for more efficient exploration.

The Upper Confidence Bound (UCB)  \cite{auer2002using} algorithm is based on the idea of optimism in the face of uncertainty. It calculates an upper confidence bound for the mean reward of each arm, which is a high probability overestimate of the value. Later, \cite{kaufmann2012bayesian} introduced BayesUCB, which uses upper quantiles of the posterior distribution to compute the UCB. Thompson Sampling (TS) \cite{thompson1933likelihood} is a classical Bayesian approach to establish the trade-off between exploration and exploitation, which assumes the existence of a prior distribution for the learnable parameters. 
Several studies, e.g., \cite{russo2014learning,agrawal2012analysis,kaufmann2012thompson}, demonstrate the performance of TS in different problem settings. Other works focus on TS in a \emph{combinatorial} setting with the objective of finding shortest paths \cite{wang2018thompson} and bottlenecks \cite{aakerblom2022online} in networks (though these works do not consider contextual information).

In the presence of additional information for arms, also called \textit{contextual} information, \emph{contextual bandit} models have been introduced, where the reward for each time step depends on the context and the played action at that time. Assuming that the expected reward is linear with the arm's context, \cite{auer2002using} proposed the contextual bandit algorithm LinRel. Later, \cite{li2010contextual} improved LinRel by using a different form of regularization and introduced the LinUCB algorithm. However, the assumption of linear rewards is not always true, especially in real-world applications. 
Therefore, different approaches have been proposed \cite{riquelme2018deep,zhang2020neural,zhou2020neural,ban2021ee} using the representation potential of deep neural networks to learn the underlying reward function. \cite{zhou2020neural} trained a neural network to learn the expected reward function and compute the upper confidence bound with respect to the expected reward conditional on the context. Their algorithm, called NeuralUCB, guarantees a near-optimal regret bound.

In this paper, we formulate the bottleneck identification task as a \textit{Contextual Combinatorial Semi-Bandit} problem and propose an online learning approach for addressing it. In this framework, we study $\epsilon_t$-greedy, LinUCB, BayesUCB, and TS under contextual combinatorial semi-bandit conditions. We also propose an extension of the neural network approach, NeuralUCB \cite{zhou2020neural}, for a combinatorial environment.
Then, we evaluate our approach on a real-world application, namely road networks. 

\section{Related Work}\label{sec:RelatedWork}

Here, we outline some of the most important related works for the topics discussed in this paper.

\subsection{Contextual Bandits}
Some assumptions are commonly made in contextual bandit problems regarding the expected reward function, inspired by the relation between the expected rewards of actions and their corresponding observed feature vectors.
Assuming the reward function is linear with the context,  \cite{Abbasi-Yadkori2011,li2010contextual,dani2008stochastic} estimated the parameters of the function using ridge regression. 
Other studies \cite{qin2014contextual,zhang2017contextual} proposed combinatorial MAB algorithms based on LinUCB with application-specific objectives. \cite{agrawal2013thompson,Abeille2018} studied variants of the Thompson Sampling algorithm with linear arm rewards, and derived regret bounds of $\tilde{\mathcal{O}}(\sqrt{T})$, where $T$ is the horizon.

To relax the linearity assumption, \cite{filippi2010parametric} proposed an approach inspired by UCB, where they generalized the reward function to a composition of a linear function and a link function.
\cite{valko2013finite} proposed KernelUCB, a kernelized UCB algorithm, for contextual bandit problems, where the algorithm finds upper confidence bounds on the expected rewards of each action using the dualization of regularized linear regression in the reproducing kernel Hilbert space (RKHS). They proved a cumulative regret bound of  $\tilde{\mathcal{O}}(\sqrt{T \tilde{d}})$, where $\tilde{d}$ is an \textit{effective dimension} upper bounded by the dimension of the contexts.

\subsection{Neural Bandits}
Inspired by the recent advances of Deep Neural Networks (DNN) in reinforcement learning, \cite{riquelme2018deep} proposed the NeuralLinear algorithm, using an $L$-layer DNN to obtain a feature map of arms' context in a lower dimension. Then they used Bayesian linear regression in the last layer of the DNN, on top of the feature map, and used Thompson Sampling to decide which arm to play.
\cite{zhou2020neural} proposed NeuralUCB, using DNN to learn the arms' expected reward functions and the UCB approach as exploration policy, with a regret bound of  $\tilde{\mathcal{O}}(\tilde{d}\sqrt{T} )$.

Unlike NeuralLinear, NeuralUCB uses the entire DNN to learn the representation of arms' context and subsequently uses the gradient of the DNN function to obtain the upper confidence bound. NeuralUCB is able to deal with unknown reward functions, as long as they are bounded.
Inspired by NeuralUCB, \cite{ban2021ee} proposed a method called Exploration Exploitation Neural Networks (EE-NET), with a regret bound of $\mathcal{O}(\sqrt{T \log T})$. EE-NET uses neural networks to estimate three functions: i) rewards for each arm, ii) potential benefits of exploring each arm, and iii) a decision-maker function to select an arm.
\cite{zhang2020neural} proposed an algorithm called Neural Thompson Sampling. To select an arm, the algorithm samples an expected reward from the reward posterior distribution, the mean of which is obtained by a DNN. Their algorithm achieves a cumulative regret of  $\mathcal{O}(\sqrt{T} )$.

\section{Problem Formulation}

In this section, we first introduce the problem of identifying bottlenecks in a road network. Then, assuming that the network is not fully specified, we define the probabilistic model of the problem.

\subsection{Road Network Formulation }  \label{sec:RoadNetwork}
Consider a road network given as a graph $\mathcal{G} (\mathcal{V}, \mathcal{E}, w)$, where the nodes $\mathcal{V}$ represent intersections in the road network, and the edges $\mathcal{E}$ represent road segments. $\mathcal{G}$ is a directed graph, where each edge $e = (u,v) \in \mathcal{E}$ is the edge leading from node $u \in \mathcal{V}$ to $v \in \mathcal{V}$. A weight function $w : \mathcal{E} \rightarrow  \mathbb R^{+}$ maps edge $e \in \mathcal{E}$ to its corresponding weight value $w(e)$. 

We let the bottleneck of a single path refer to the edge on the path with the highest weight value. Consequently, given an arbitrary pair of a source node $v \in \mathcal{V}$ and a target node $u \in \mathcal{V}$, the bottleneck of all possible paths connecting $v$ and $u$, is the smallest bottleneck among all possible paths connecting $v$ to $u$.
In other words, the bottleneck of $v$ and $u$, denoted by $M(v,u,\mathcal{G})$, can be defined as the edge with the minimax distance between these two points, as obtained by Eq. \ref{eq:MM} where $p_{v,u}$ is a path connecting the source $v$ and the target node $u$, and $\mathcal{P}_{v,u}$ is the set of all paths between $v$ and $u$. Note that, to use minimax distance, the weight function $w$ is required to respect semi-metric constraints.
\begin{equation}\label{eq:MM}
    M(v,u,\mathcal{G}) =  \arg \min_{p_{v,u} \in \mathcal{P}_{v,u}} \max_{e\in p_{v,u}} w(e) 
\end{equation}
We apply a modification \cite{berman1987optimal} of Dijkstra's algorithm \cite{dijkstra1959note}  to find the path including the bottleneck edge, $M(v,u,\mathcal{G})$, in the directed graph setting.

\subsection{Probabilistic Model of the Road Network} \label{sec:ProbabilisticModel}
In our problem setting, the edge weights of the network are uncertain and stochastic. Thus, we adopt a Bayesian approach to model the edge weights and make use of prior knowledge. Given a time horizon $T$, for each edge $e \in \mathcal{E}$ and time step $t \in [T]$, we assume access to a $d$-dimensional feature vector $\bm{c}_{t,e}$ representing contextual information related to the edge at that point in time.  We further assume that the expected value of the weight is linear in $\bm{c}_{t,e}$, for some unknown parameter vector $ \bm{\theta}_e^* \in \mathbb{R}^d$.

More specifically, we assume that the weight $w_t(e)$ at time $t$, given $\bm{c}_{t,e}$, for all $e \in \mathcal{E}$, is sampled independently from a Gaussian distribution $\mathcal{N}(\bm{c}_{t,e}^{\top} \cdot \bm{\theta}_e^*,\,\varsigma^{2}_e)$, where the mean vector $\bm{\theta}^*_e$ is itself sampled from a prior distribution $ \mathcal{N}(\bm{\mu}_{0,e},\,\Sigma_{0,e})$. For convenience, like other works \cite{russo2014learning,agrawal2013thompson}, we assume $\varsigma_e^2$ is known. With a Gaussian prior and likelihood, due to conjugacy, the posterior has the same parametric form as the prior and 
the update of the parameters of the Gaussian posterior is derived in a closed form.

\section{The Contextual Combinatorial Semi-bandit Framework}  \label{sec:framework}
We formulate the network bottleneck identification task as a contextual combinatorial semi-bandit problem. We have a set of base arms $\mathcal{A} = \{1, \dots, K\}$, each of which corresponds to an edge in the graph. Furthermore, we consider a set of super-arms $\mathcal{I} \subseteq 2^\mathcal{A}$, corresponding to the set of all paths between a specified source node and target node. In each time step $t \in [T]$, for each base arm $a \in \mathcal{A}$, the environment reveals a contextual feature vector $\bm{c}_{t,a}$ to the agent. Subsequently, the agent selects a super-arm $\mathcal{S}_t \in \mathcal{I}$. 

For each base arm $a \in \mathcal{S}_t$, where $\mathcal{S}_t $ is a super-arm $\mathcal{S}_t \in \mathcal{I}$  at time $t$ , the environment reveals a loss sampled from $\mathcal{N}(\bm{c}_{t,a}^{\top} \cdot \bm{\theta}_a^*, \varsigma^2_a)$ to the agent, i.e., semi-bandit feedback. The cost for a super-arm $c(\mathcal{S}_t)$ is defined as the maximum of losses of any base arm $a \in \mathcal{S}_t$, and the expected cost of a super-arm $\mathcal{S}_t$  is $f_{\bm{\theta}}(\mathcal{S}_t) := \E{\max_{a \in \mathcal{S}_t }X_a}$, where $ X_a \sim \mathcal{N}(\bm{c}_{t,a}^{\top} \cdot \bm{\theta}_a, \varsigma^2_a)  $ is the loss for arm $a$. Then, the  objective of the agent, in each time step $t$, is to find a super-arm which minimizes the expected maximum base arm loss, conditioned on the contextual information.

The mentioned objective is computationally intractable, since a super-arm may contain many base arms. Therefore, we use an approximate objective \cite{aakerblom2022online}, altered to find a super-arm $\mathcal{S}_t \in \mathcal{I}$ minimizing the maximum expected base arm loss, given the contextual information.
This can be formulated as a regret minimization task over the time horizon $T$, where we define the expected cumulative regret as follows:

\begin{equation}
    \text{Regret}(T) =\mathbb{E}  \left[  \left(\sum_{t=1}^{T} \max_{a \in \mathcal{S}_t} \bm{c}_{t, a}^{\top} \cdot \bm{\theta}^*_a\right) - \left(\sum_{t=1}^{T} \max_{a \in \mathcal{S}_t^*} \bm{c}_{t, a}^{\top} \cdot \bm{\theta}^*_a\right) \right],  \label{eq:exact_regret}
\end{equation}
where  $\mathcal{S}_t^* =  \arg \min_{\mathcal{S} \in \mathcal{I}} \max_{a \in \mathcal{S}} \bm{c}_{t, a}^{\top} \cdot \bm{\theta}^*_a $.

\subsection{The Framework}

We adapt and investigate several contextual combinatorial semi-bandit methods for the bottleneck identification problem. 
Algorithm \ref{alg:general} describes our general framework for this task. Given a graph $\mathcal{G}(\mathcal{V}, \mathcal{E})$, a source node $v \in \mathcal{V}$ and a target node $u \in \mathcal{V}$, the agent in each time step $t\in [T]$, for each edge $e \in \mathcal{E}$, observes the context vector $\bm{c}_{t,e}$. Each agent keeps and updates an internal representation of the graph $\mathcal{G}$, with weights $\hat{w}_t(e)$ assigned using the context $\bm{c}_{t,e}$ to induce exploration.
Having observed $\bm{c}_{t,e}$, the agent subsequently updates the internal edge weight $\hat{w}_t(e)$ according to $\bm{c}_{t,a}$ and algorithm-specific parameter estimate $\hat{\bm{\theta}}_{t,e}$.
Different algorithms have different approaches to update $\hat{w}_t(e)$, which we discuss later in this section. In \cref{alg:genreal:line:dijkstra}, the agent finds a path $p_{v,u}$ containing the bottleneck $M(v,u,\mathcal{G}(\mathcal{V}, \mathcal{E}, \hat{w}_t))$ based on the modification of Dijkstra's algorithm. 
In  \crefrange{alg:genreal:line:For}{alg:genreal:line:update}  the agent, for each played edge $e$ in the path, observes the stochastic edge weight (arm loss) $w_t(e)$ and then updates the parameters $\hat{\theta}_{t,e}$, and internal weights $\hat{w}_t (e)$ accordingly.
 
\begin{algorithm}[b]
\caption{General Framework}
\textbf{Input:} $\mathcal{G}(\mathcal{V},\mathcal{E}), v,u$ 
\label{alg:general}
\begin{algorithmic}[1] %[1] enables line numbers
\For{$t \leftarrow 1, \dots, T$}
  \For{$e \in \mathcal{E}$}
    \State Observe the edge context $\bm{c}_{t,e}$ 
    \State Assign algorithm-specific edge weight $\hat{w}_t (e)$ based on $\hat{\bm{\theta}}_{t,e}$ and  $\bm{c}_{t,e}$ 
  \EndFor
  \State Obtain path $p_{v,u}$  that contains $M(v,u,\mathcal{G}(\mathcal{V}, \mathcal{E}, \hat{w}_t))$ \label{alg:genreal:line:dijkstra}
  \For{ $e\in p_{v,u}$}  \label{alg:genreal:line:For}
    \State Observe $w_t (e)$ by traversing edge $e$ \label{alg:genreal:line:traversing}
    \State Update the parameter estimate $\hat{\bm{\theta}}_{t,e}$\label{alg:genreal:line:update}
  \EndFor
\EndFor
\end{algorithmic}
\end{algorithm}

\subsection{Adapted Algorithms}
We adapt several methods to our contextual combinatorial semi-bandit framework.

-\textbf{ Thompson Sampling:} In  Thompson Sampling, we have a prior distribution $\mathcal{N}(\bm{\mu}_{0,e},\,\Sigma_{0,e})$ for $\bm{\theta}^*_e$ where $\bm{\mu}_{0,e} \in \mathbb{R}^{d\times 1}$ is the prior mean vector and $\Sigma_{0,e} \in \mathbb{R}^{d\times d}$ is the prior covariance matrix. 
The prior is a conjugate prior to the model, so the posterior is also Gaussian, with efficient closed form expressions available for parameter updates. 
Algorithm \ref{alg:TS} describes the TS agent. In  \cref{alg:TS:line:sampling}, the agent samples $\hat{\bm{\theta}}_{t,e}$ from the posterior distribution $\mathcal{N}(\bm{\mu}_{t-1,e},\Sigma_{t-1,e}) $. Then, it updates $\hat{w}_t (e)$ with the expected weight given the sampled  $\hat{\bm{\theta}}_{t,e}$. The TS agent updates the posterior parameters by UPDATE-PARAMS procedure shown in Algorithm \ref{alg:update}.

 \begin{algorithm}[ht]
\caption{Thompson Sampling for bottleneck identification}
\textbf{Input:} $\mathcal{G}(\mathcal{V},\mathcal{E}), v,u, \bm{\mu}_{0},\Sigma_{0}, \varsigma^2$
\label{alg:TS}
\begin{algorithmic}[1] %[1] enables line numbers
\For{$t \leftarrow 1, \dots, T$}
\For{$e \in \mathcal{E}$}
\State Observe context $\bm{c}_{t,e}$ 
\State $\hat{\bm{\theta}}_{t,e} \gets$ Sample from posterior $\mathcal{N}(\bm{\mu}_{t-1,e},\,\Sigma_{t-1,e})$ \label{alg:TS:line:sampling}
\State $\hat{w}_t (e) \gets \bm{c}_{t,e}^{\top} \cdot \hat{\bm{\theta}}_{t,e}$
\EndFor
\State $p_{v,u} \gets$ Obtain a path that contains $M(v,u,\mathcal{G} (\mathcal{V}, \mathcal{E}, \hat{w}_t))$
\For { $e \in p_{v,u}$}
\State Observe $w_t (e)$ by traversing edge $e$ 
\State $ \bm{\mu}_{t,e}, \Sigma_{t,e} \gets$UPDATE-PARAMS($\bm{c}_{t,e}, w_t (e), \bm{\mu}_{t-1,e}, \Sigma_{t-1,e}, \varsigma_e^2$) 
\EndFor
\EndFor
\end{algorithmic}
\end{algorithm}

-\textbf{ BayesUCB:}
We adapt the combinatorial BayesUCB method in \cite{aakerblom2020online} for the contextual setting, where the agent uses the upper quantiles of the posterior distributions over expected arm rewards to determine the level of exploration.
The quantile function of a distribution $\rho$, denoted by $\mathcal{Q}(\nu, \rho)$, is defined so that $\mathbb{P}_\rho(X \leq \mathcal{Q}(\nu,\rho)) = \nu$. In Algorithm \ref{alg:BayesUCB}, describing BayesUCB, lower quantiles of the expected edge weights given the context are used in line \ref{alg:BayesUCB:line:quantile}, since the stochastic edge weights may be interpreted as negative rewards. The BayesUCB agent makes use of UPDATE-PARAMS procedure.

\begin{algorithm}[b]
\caption{Bayesian posterior parameter update procedure }
\label{alg:update}
\begin{algorithmic}[1] %[1] enables line numbers
\State procedure UPDATE-PARAMS($\bm{c}_{t,e}, w_t (e), \bm{\mu}_{t-1,e} ,\Sigma_{ t-1,e} , \varsigma_e^2$ )
\State \hspace{\algorithmicindent}$\Sigma_{t,e} \leftarrow \left(\Sigma_{t-1,e}^{-1} + \frac{  1}{\varsigma^2_e} \bm{c}_{t,e} \bm{c}_{t,e}^{\top}\right )^{-1} $
\State \hspace{\algorithmicindent}$ \bm{\mu}_{t,e} \leftarrow \Sigma_{t,e} \left(\Sigma_{t-1,e}^{-1} \bm{\mu}_{t-1,e} + \frac{ 1}{\varsigma^2_e } w_{t} (e) \bm{c}_{t,e}\right)$
\end{algorithmic}
\end{algorithm}

\begin{algorithm}[ht]
\caption{BayesUCB for bottleneck identification}
\textbf{Input:} $\mathcal{G}(\mathcal{V},\mathcal{E}), v,u,\bm{\mu}_{0},\Sigma_{0}, \varsigma^2$
\label{alg:BayesUCB}
\begin{algorithmic}[1] %[1] enables line numbers
\For{$t \leftarrow 1, \dots, T$}
\For{$e \in \mathcal{E}$}
\State Observe context $\bm{c}_{t,e}$ 
\State $\hat{w}_t (e) \gets \bm{c}_{t,e}^{\top} \bm{\mu}_{t-1,e}  - \mathcal{Q}(1 - \frac{1}{t}, \mathcal{N}(0,1)) \sqrt{\bm{c}_{t,e}^{\top}\Sigma_{t-1,e} \bm{c}_{t,e}}$ \label{alg:BayesUCB:line:quantile}
\EndFor
\State $p_{v,u} \gets$ Obtain a path that contains $M(v,u,\mathcal{G}(\mathcal{V}, \mathcal{E}, \hat{w}_t))$
\For{ $e \in p_{v,u}$}
\State Observe $w_t (e)$ by traversing edge $e$ 
\State $ \bm{\mu}_{t,e}, \Sigma_{t,e} \gets$UPDATE-PARAMS($\bm{c}_{t,e}, w_t (e), \bm{\mu}_{t-1,e}, \Sigma_{t-1,e}, \varsigma_e^2$)
\EndFor
\EndFor
\end{algorithmic}
\end{algorithm}

-\textbf{ LinUCB:} 
The LinUCB agent, described in Algorithm \ref{alg:LinUCB}, introduces the parameters $\bm{A}_{t,e} \in \mathbb{R}^{d\times d} $ and $\bm{b}_{t,e} \in \mathbb{R}^{d\times 1}$  for each edge $e \in \mathcal{E}$. 
To find the UCB value for each edge, first, the algorithm obtains the ridge regression estimate $\hat{\bm{\theta}}_{t,e} \gets  \bm{A}_{t-1,e}^{-1} \bm{b}_{t-1,e}$. Then, using the exploration factor $\alpha$ to determine the degree of exploration, the agent calculates the current internal edge weight values. Since our problem is formulated as a loss minimization, we negate the exploration term. The procedure for updating the parameters in LinUCB follows  \cref{alg:linUCB:line:updateA,alg:linUCB:line:updateB}.  

\begin{algorithm}[ht]
\caption{LinUCB for bottleneck identification}\label{alg:LinUCB}
\textbf{Input:} $\mathcal{G}(\mathcal{V},\mathcal{E}), v,u,  \bm{A}_{0,e} ,\bm{b}_{0,e}, \alpha $ , $\varsigma_e^2$
\begin{algorithmic}[1] %[1] enables line numbers
\For{$t \leftarrow 1, \dots, T$}
\For{$e \in \mathcal{E}$}
\State Observe context $\bm{c}_{t,e}$ 
\State $\hat{\bm{\theta}}_{t,e} \gets \bm{A}_{t-1,e}^{-1} \bm{b}_{t-1,e}$
\State $\hat{w}_t (e) \gets \hat{\bm{\theta}}_{t,e}^{\top} \bm{c}_{t,e} - \alpha \sqrt{\bm{c}_{t,e}^{\top} \bm{A}_{t-1,e}^{-1} \bm{c}_{t,e} }$
\EndFor
\State $p_{v,u} \gets$ Obtain a path that contains $M(v,u,\mathcal{G}(\mathcal{V}, \mathcal{E}, \hat{w}_t ))$
\For{ $e \in p_{v,u}$}
\State Observe $w_t (e)$ by traversing edge $e$ 
\State $\bm{A}_{t,e} \gets \bm{A}_{t-1,e} +  \frac{1}{\varsigma^2_e} \bm{c}_{t,e} \bm{c}_{t,e}^{\top} $ \label{alg:linUCB:line:updateA} 
\State $\bm{b}_{t,e} \gets \bm{b}_{t-1,e} + \frac{1}{\varsigma^2_e} w_{t} (e) \bm{c}_{t,e}$\label{alg:linUCB:line:updateB}
\EndFor
\EndFor
\end{algorithmic}
\end{algorithm}

-\textbf{ NeuralUCB:} We extend the NeuralUCB algorithm in \cite{zhou2020neural} for the combinatorial setting as described by Algorithm \ref{alg:nnUCB}. We assume a reward function $h(\bm{c}_{t,e}; \bm{\theta}_t)$, where $\bm{\theta}_t$ indicates the learnable parameters, at time $t$, of a neural network with $L$ layers, and $\rho$ is the number of learnable parameters. Then, the gradient $ g(\bm{c}_{t,e}; \bm{\theta}_t) = \nabla_{ \bm{\theta}_t}  h(\bm{c}_{t,e}; \bm{\theta}_t)$ of the neural network function is used to calculate the arms' upper confidence bounds. In \cref{alg:nnUCB:line:edgewieght}, the agent updates $\hat{w}_t (e) $ with respect to the reward function $h$, its gradient $g$, a scaling factor $\gamma_t$, and the number of neurons in a hidden layer $m$ (assuming equal width for all $L$ layers, for simplicity). The algorithm trains the network in \crefrange{alg:nnUCB:line:train1}{alg:nnUCB:line:train2}, where $J$ is the number of gradient descent steps, $\eta$ is the step size, and $\nabla \mathcal{L}(\bm{\theta}^j)$ for $ j\in[J]$  is the gradient of the loss function, where the loss function is defined as $ \mathcal{L}(\bm{\theta}^j) = \sum_{e \in p_{v,u}}( h(\bm{c}_{t,e}; \bm{\theta}^j) - w_{t} (e))^2 +  \lambda  \lVert \bm{\theta}^j  \rVert_2 $  at time $t$, with regularization parameter $\lambda$. In \cref{alg:nnUCB:line:updateGamma}, we update $\gamma_{t-1}$ according to the update formula of the scaling factor in \cite{zhou2020neural}.

\begin{algorithm}[ht]
\caption{NeuralUCB for bottleneck identification}\label{alg:nnUCB}
\textbf{Input:} $\mathcal{G}(\mathcal{V},\mathcal{E}), v,u, \bm{Z}_{0}, \bm{\theta}_0, \rho, \lambda, m, \gamma_0, J, \eta  $
\begin{algorithmic}[1] 
\For{$t \leftarrow 1, \dots, T$}
    \For{$e \in \mathcal{E}$}
    \State Observe context $\bm{c}_{t,e}$ 
    \State $\hat{w}_t (e) \gets  h(\bm{c}_{t,e}; \bm{\theta}_{t-1}) - \gamma_{t-1} \sqrt{ g(\bm{c}_{t,e}; \bm{\theta}_{t-1})^{\top} \bm{Z}_{t-1}^{-1}  g(\bm{c}_{t,e}; \bm{\theta}_{t-1})/m}$ \label{alg:nnUCB:line:edgewieght}
    \EndFor
    \State $p_{v,u} \gets$ Obtain a path that contains $M(v,u,\mathcal{G}(\mathcal{V}, \mathcal{E}, \hat{w}_t )$
    \State $\bm{Z}_{t} = \bm{Z}_{t-1} + \sum_{e \in p_{v,u}} g(\bm{c}_{t,e}; \bm{\theta}_{t-1}) g(\bm{c}_{t,e}; \bm{\theta}_{t-1})^{\top}/m $
    \State $\bm{\theta}^{0} \gets \bm{\theta}_{t-1}$\label{alg:nnUCB:line:train1}
    \For{$j= 0, \dots, J-1$} 
        \State  $ \bm{\theta}^{j+1} = \bm{\theta}^{j}- \eta \nabla \mathcal{L}(\bm{\theta}^j)$ \label{alg:nnUCB:line:gradient}
    \EndFor
    \State $  \bm{\theta}_{t} \gets \bm{\theta}^{J}$ \label{alg:nnUCB:line:train2}
    \State  $\gamma_t \gets $Update $\gamma_{t-1}$ \label{alg:nnUCB:line:updateGamma}
    \EndFor
\end{algorithmic}
\end{algorithm}

-\textbf{ $\epsilon_t$-greedy:} We adapt the $\epsilon_t$-greedy algorithm in the supplementary material of \cite{chen2013combinatorial} to our problem setting, with $\epsilon_t = t^{-1/2}$.  
The algorithm estimates $\hat{\bm{\theta}}_{t,e}$ as the mode of the posterior distribution $\mathcal{N}(\bm{\mu}_{t,e},\Sigma_{t,e})$, and then updates the internal edge weights by  $\bm{c}_{t,e}^{\top}\hat{\bm{\theta}}_{t,e}$. Then, with probability $1-\epsilon_t$, the agent greedily selects the path based on the current estimates, and with probability $\epsilon_t$, it chooses an arbitrary path containing a uniformly random edge. The update of the parameters follows Algorithm \ref{alg:update}.

\section{Experiments}
In this section, we evaluate our framework in a real-world  road network application. We use the simulation framework in \cite{russo2018tutorial}, adapt it to the road networks,  and extend it to contextual combinatorial settings.

\begin{figure*}[ht]
     \centering
     \begin{subfigure}[b]{0.49\textwidth}
         \centering
         \includegraphics[width=\textwidth]{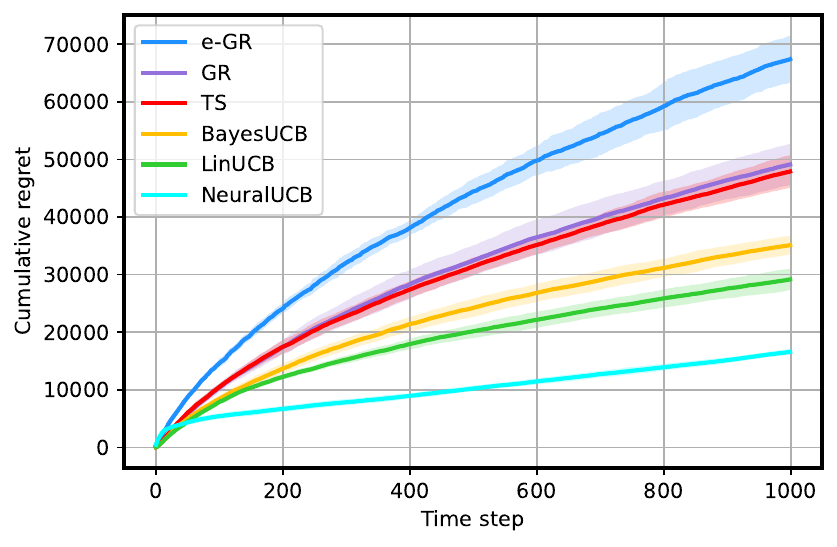}
         \caption{Cumulative regret for each time step}
         \label{fig:lux:cumulative}
     \end{subfigure}
     \begin{subfigure}[b]{0.49\textwidth}
         \centering
         \includegraphics[width=\textwidth]{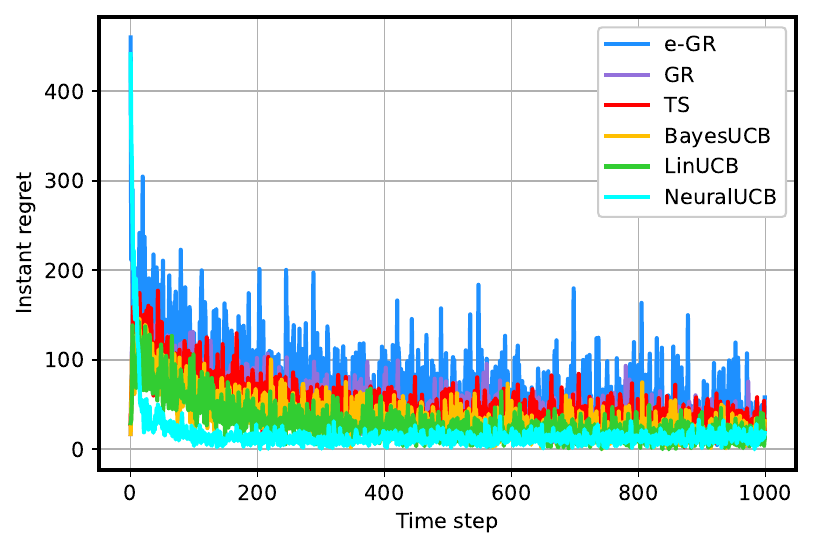}
         \caption{Instant regret for each time step}
         \label{fig:lux:Instant}
     \end{subfigure} 
    \begin{subfigure}[b]{0.49\textwidth}
         \centering
         \includegraphics[trim={0 0 0 0.5cm},clip, width=\textwidth]{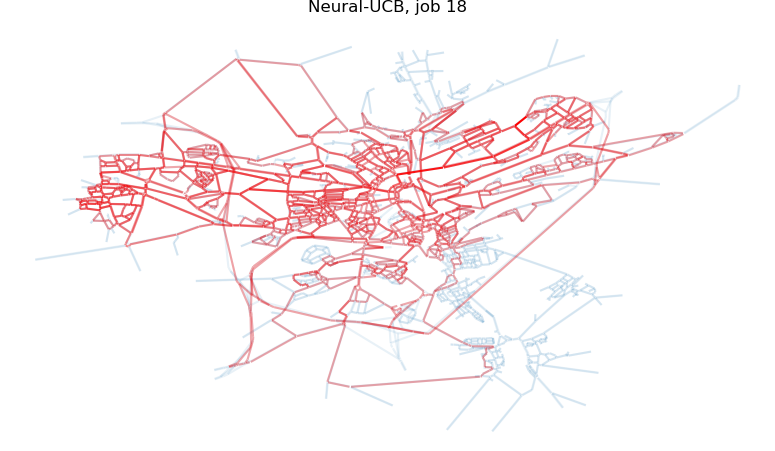}
         \caption{Explored paths by NeuralUCB }
         \label{fig:lux:road_nnUCB}
     \end{subfigure}   
     \begin{subfigure}[b]{0.49\textwidth}
         \centering
         \includegraphics[trim={0 0 0 0.5cm},clip, width=\textwidth]{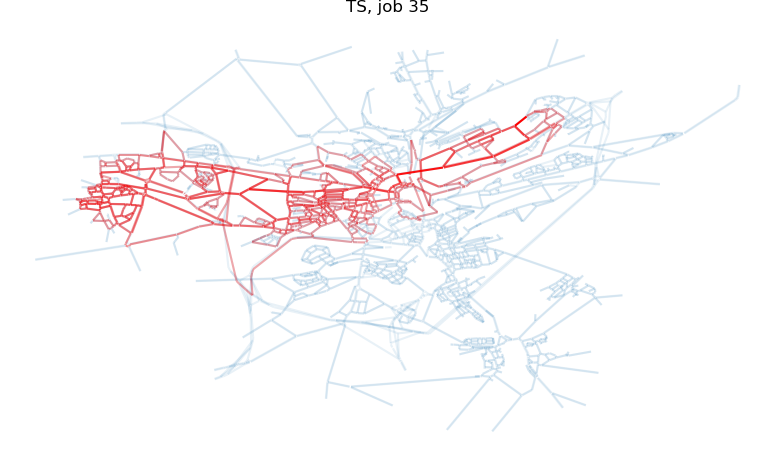}
         \caption{Explored paths by TS }
         \label{fig:lux:road_TS}
     \end{subfigure}

        \caption{Average of the cumulative with standard error regions (a) and instant (b) regret of TS, BayesUCB, LinUCB, NeuralUCB, $\epsilon_t$-greedy (e-GR) and Greedy (GR) at each time step over 10 runs. Visualized exploration of the road segments in the Luxembourg road network, by NeuralUCB (c) and TS (d). }
        \label{fig:result}
\end{figure*}

\subsection{Road Network Dataset}

To evaluate our method, we use map data for the road network around Luxembourg City from the Luxembourg SUMO Traffic (LuST) Scenario \cite{codeca2017luxembourg}, originally derived from OpenStreetMap (OSM) \cite{OpenStreetMap}. The road network contains $\sim2,000$ intersections and $\sim8,000$ road segments. For each intersection, there is information like IDs and coordinates. For each road segment, we are provided with information such as the IDs of the two intersections connected by the road segment, as well as the length of the road segment. Furthermore, we add altitude information \cite{farr2000shuttle} to each intersection of the road network.

\subsection{Experimental Setup}
For each edge $e \in \mathcal{E}$, the vector $\bm{c}_{t,e} \in \mathbb{R}^d$ contains the contextual information, with $d=12$. This vector contains two types of information. For $i \in [1,6]$ each element $c_{t,e}^{(i)}$ is sampled independently from a standard Gaussian distribution, such that $c_{t,e}^{(i)} \sim \mathcal{N}(0, 1)$. The second half of the vector, where $i \in [7, 12]$, contains edge-specific spatial coordinate information. Unlike other agents, which model and learn edge-specific parameters, NeuralUCB maintains a single model for all edges. Therefore, features that can be used to distinguish different edges are crucial for this agent. For these features, we utilize the longitude, latitude and altitude of the edge start and end vertices.
BayesUCB and TS make use of the prior distribution $ \mathcal{N}(\bm{\mu}_{0,e},\,\Sigma_{0,e})$, where $ \bm{\mu}_{0,e} $ is a $d$-dimensional vector with every element assigned as $100$, and $\Sigma_{0,e} = 100 \cdot\mathbb{I}_d $ . 
For the LinUCB agent,  we set the exploration factor $\alpha = 4$. While LinUCB does not require a prior distribution, we utilize $ \bm{\mu}_{0,e} $ and $ \Sigma_{0,e}$ to initialize it as similar to BayesUCB as possible. Thus, we set  $\bm{A}_{0,e} = \Sigma_{0,e}^{-1}$, and $\bm{b}_{0,e} = \bm{A}_{0,e} \bm{\mu}_{0,e} $. 
The NeuralUCB agent utilizes a $2$-layer neural network with $10$ nodes per layer and ReLU as the activation function. We initialize $\bm{Z}_{0}= 0.01 \cdot \mathbb{I}_\rho$, where $\rho =d(m+1)+m(m+1)+m+1 = 251$, and $ \bm{\theta}_0$ by drawing samples from a uniform distribution$(-1/\sqrt{m},\ 1/\sqrt{m})$.
We set the regularization parameter $ \lambda=0.001$, scaling factor $\gamma_0 = 0.01 $, number of gradient descent steps $J = 10$, and gradient step size $\eta = 0.001$. We update $\gamma_t$ with the formulation provided in \cite{github}. We fix the horizon as $T = 1000$, and repeat all experiments for $10$ runs, where the true underlying parameters $\bm{\theta}_e^*$ for each edge $e \in \mathcal{E}$ is sampled independently from the prior $ \mathcal{N}(\bm{\mu}_{0,e},\,\Sigma_{0,e})$, in the beginning of each experiment run.

\subsection{Experimental Results}

 \Cref{fig:lux:cumulative} shows the cumulative regret of the agents at each time step, averaged over $10$ runs.
NeuralUCB yields the best performance and stabilizes faster compared to the others. It also has the smallest error region among the agents.  \Cref{fig:lux:road_nnUCB} shows the road network of Luxembourg, where the paths explored by NeuralUCB are colored red, with opacity indicating the amount of exploration of each road segment. The figure shows that the NeuralUCB agent has explored a majority of the road segments at least once.

After that, LinUCB and BayesUCB compete closely, in terms of performance. LinUCB performs slightly better and has a narrower error region compared to BayesUCB. Thompson Sampling performs slightly worse than the others and is marginally better than Greedy. This might be because, with our objective (explained in \cref{sec:framework}), TS prefers paths with fewer segments. This preference is shown in figure \ref{fig:lux:road_TS}. The error region of TS is similar to that of BayesUCB. $\epsilon_t$-greedy has the widest error region and the highest final cumulative regret. The fairly good performance of the Greedy method (compared to TS and $\epsilon_t$-greedy) might be explained by the variability of the context in our experimental setup (i.e., half of the features are sampled from Gaussian distributions). This variability may induce some explorative behavior in the method, but is possibly not present in all problem settings.
 \Cref{fig:lux:Instant} illustrates the average \emph{instant} regret of each agent and time step, which decreases as we progress in time, and it is again apparent that NeuralUCB yields the best performance. 

\section{Conclusion}
We developed a unified contextual combinatorial bandit framework for network bottleneck identification which takes into account the available side information. We introduced NeuralUCB in the combinatorial setting.
We employed a minimax concept to find the path containing the bottleneck edge.
In addition to NeuralUCB, we adapted Thompson Sampling, BayesUCB, LinUCB and $\epsilon_t$-greedy to our framework. We evaluated its performance on a real-world road network application and demonstrated that NeuralUCB outperforms the other agents in our application.

\section*{Acknowledgments}
This work is partially funded by the Strategic Vehicle Research and Innovation Programme (FFI) of Sweden, through the project EENE (reference number: 2018-01937).

\bibliography{main}

\end{document}